\pdfoutput=1
\documentclass[11pt]{article}

\usepackage{booktabs}
\usepackage{EMNLP2023}
\usepackage{amsmath} 
\usepackage{xcolor}
\usepackage[most]{tcolorbox}

\usepackage{times}
\usepackage{amssymb}
\usepackage{latexsym}
\usepackage{pgfplots}
\usepackage{multirow}
\usepackage[table,xcdraw]{xcolor}
\pgfplotsset{compat=1.18}

\newcommand{\myparagraph}[1]{\smallskip\noindent{\bf {#1}.}~}
\usepackage[T1]{fontenc}
\usepackage[utf8]{inputenc}

\usepackage{microtype}

\usepackage{inconsolata}

\usepackage{caption}

\usepackage{pgfplots}
\pgfplotsset{compat=1.16}

\definecolor{c3}{RGB}{155,191,138}
\definecolor{c2}{RGB}{130,175,218}
\definecolor{c5}{RGB}{247,147,89}
\definecolor{c1}{RGB}{154,132,191}
\definecolor{c4}{RGB}{255,190,122}
\definecolor{c6}{RGB}{200,36,35}

\title{The LLM Already Knows: Estimating LLM-Perceived Question Difficulty via Hidden Representations}

\author{
Yubo Zhu$^{1,2}$\thanks{~Equal contribution.}\thanks{~~This work was done during an internship at Shanghai Artificial Intelligence Laboratory, supervised by Dongrui Liu.}, Dongrui Liu$^{2}$\footnotemark[1], Zecheng Lin$^{2,3}$\footnotemark[2],
Wei Tong$^{1}$\thanks{~Corresponding author.}, Sheng Zhong$^{1}$,
Jing Shao$^{2}$\footnotemark[3] \\
$^{1}$ State Key Laboratory for Novel Software Technology, Nanjing University \\
$^{2}$ Shanghai Artificial Intelligence Laboratory \\
$^{3}$ Xidian University \\
\texttt{zyb@smail.nju.edu.cn} \quad
\texttt{\{wtong, zhongsheng\}@nju.edu.cn} \\ \texttt{zechenglin@stu.xdu.edu.cn}\quad
\texttt{\{liudongrui, shaojing\}@pjlab.org.cn}
}

\newtcolorbox[auto counter, number within=section]{PromptBox}[2][]{%
  float*=t,
  width=\textwidth,
  colback=gray!5!white,
  colframe=gray!50!black,
  title=Box~\thetcbcounter: #2,
  label=#1,
  fonttitle=\bfseries,
  before skip=10pt,
  after skip=10pt,
  boxrule=0.5pt,
  arc=2mm,
  outer arc=1mm,
  left=5pt,
  right=5pt,
  top=5pt,
  bottom=5pt,
  fontupper=\small,
}

\begin{document}
\maketitle
\begin{abstract}
Estimating the difficulty of input questions as perceived by large language models (LLMs) is essential for accurate performance evaluation and adaptive inference. Existing methods typically rely on repeated response sampling, auxiliary models, or fine-tuning the target model itself, which may incur substantial computational costs or compromise generality. In this paper, we propose a novel approach for difficulty estimation that leverages only the hidden representations produced by the target LLM. We model the token-level generation process as a Markov chain and define a value function to estimate the expected output quality given any hidden state.
This allows for efficient and accurate difficulty estimation based solely on the initial hidden state, without generating any output tokens. Extensive experiments across both textual and multimodal tasks demonstrate that our method consistently outperforms existing baselines in difficulty estimation. Moreover, we apply our difficulty estimates to guide adaptive reasoning strategies, including Self-Consistency, Best-of-N, and Self-Refine, achieving higher inference efficiency with fewer generated tokens.

\end{abstract}

\section{Introduction}
\begin{figure}[!t]
    \small
    \centering
    \includegraphics[width=0.45\textwidth]{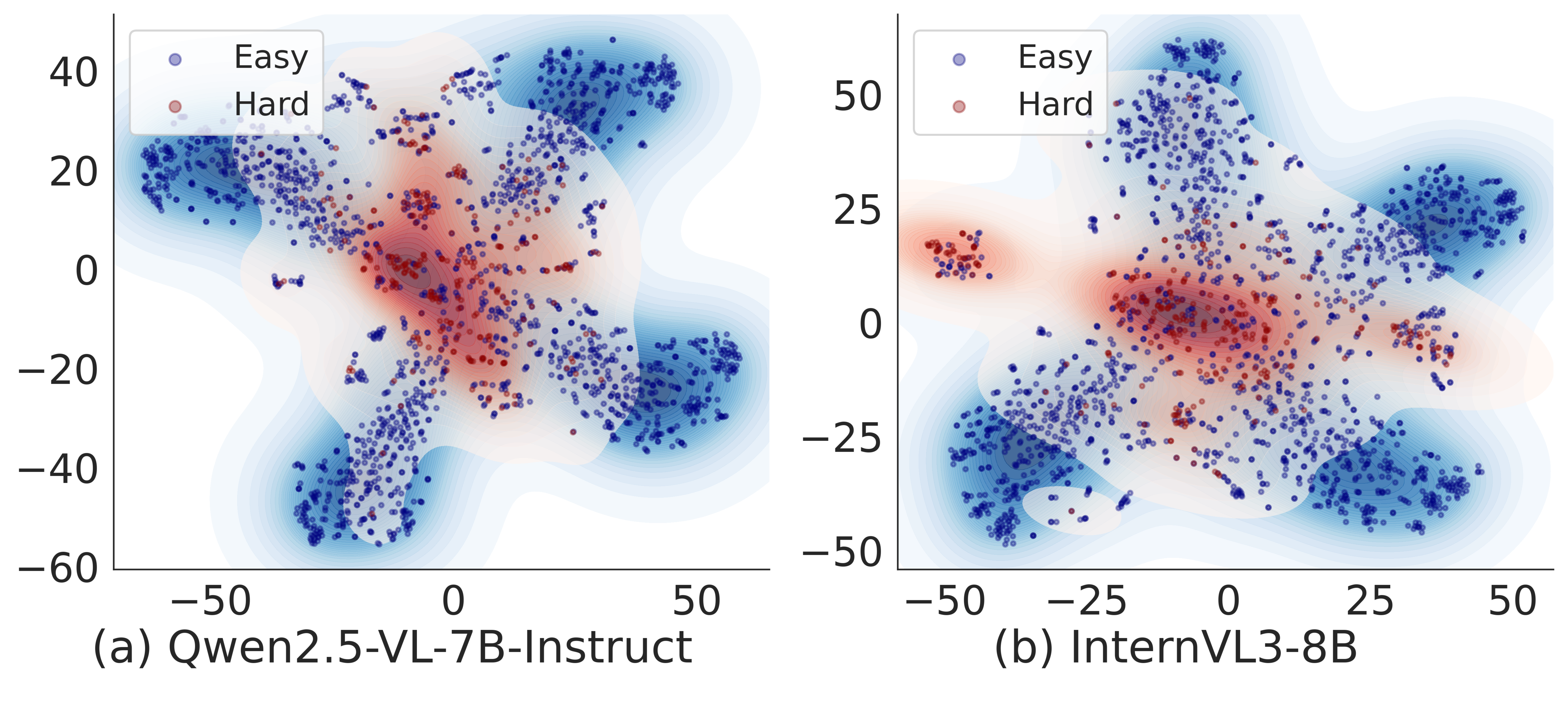}
    \caption{t-SNE visualization of last layer hidden representations derived from input questions in a randomly sampled 50\% subset of MMBench~\cite{MMBench}. Blue and orange contours represent the distributions of hard and easy problems for the LLM, respectively, with darker regions indicating areas of higher density. Here, problems are classified as easy if the model’s outputs across three independent rollouts consistently match the ground-truth answers, and as hard otherwise.}
    \label{fig:problem}
\end{figure}
%\vspace{-5pt}

As large language models (LLMs) grow more capable, accurately estimating question difficulty is becoming increasingly critical. Precise difficulty estimation not only supports difficulty-aware evaluation~\cite{ding2024easy2hard,gao2024omni,he2025deepmath}, enabling finer‑grained assessment of model performance across varying difficulty levels, but also facilitates difficulty‑aware training~\cite{xue2025dast,tian2025deepdistill,ji2025difficulty}, improving model robustness and performance on challenging examples. Furthermore, difficulty estimation drives the dynamic adaptation of test-time inference strategies~\cite{pan2024dynathink, wang2024make, li2024escape}. By tailoring inference effort to input difficulty, LLMs can allocate resources more efficiently.

%=== With the growing capabilities of multi-modal large language models (MLLMs), accurately estimating input question difficulty has become critical. More accurate difficulty estimation facilitates difficulty-aware training~\cite{xue2025dast, tian2025deepdistill, ji2025difficulty}, helping models generalize to complex inputs better. In addition, it supports difficulty-aware evaluation~\cite{ding2024easy2hard, gao2024omni, he2025deepmath}, allowing finer-grained assessment of model performance across varying difficulty levels. Furthermore, difficulty estimation drives the dynamic adaptation of test-time inference strategies~\cite{pan2024dynathink, wang2024make, li2024escape}. By adapting inference effort to input difficulty, MLLMs can allocate resources more efficiently.

%  For example, difficulty estimation can guide repeated sampling techniques by adjusting the number of samples based on question difficulty, thereby reducing computational costs without compromising accuracy.

To this end, several attempts have been made to distinguish between easy and difficult questions. These include estimating question difficulty by measuring the consistency of outputs generated by the target LLM~\cite{lee2025semantic, pan2024dynathink, li2024escape, chen2024magicore, snell2024scaling}, fine-tuning the target LLM to predict input difficulty and dynamically adapt its strategies~\cite{huang2025efficient, manvi2024adaptive}, and employing an auxiliary LLM to assess question difficulty~\cite{chen2024magicore, cheng2025think, wang2024make}. These methods mainly rely on generated outputs and typically suffer from at least one of the following limitations: substantial computational overhead due to repeated output generation; potential degradation of the target LLM’s general capabilities, including robustness and safety~\cite{qi2024fine}; and insufficient measurement of the relationship between the target LLM’s internal state and the difficulty of the input question.

%=== To this end, prior works on distinguishing between easy and difficult questions can be roughly divided into three categories. Firstly, some methods~\cite{lee2025semantic, pan2024dynathink, li2024escape, chen2024magicore, snell2024scaling} estimate question difficulty by measuring the consistency of outputs generated by the \textcolor{blue}{target LLM}. These approaches often require generating multiple outputs at test time to assess consistency, introducing significant computational overhead. \textcolor{blue}{Second, some methods~\cite{huang2025efficient, manvi2024adaptive} fine-tune the target LLM to estimate input difficulty and dynamically adapt strategies accordingly. Although these methods can improve performance on specific reasoning tasks, they compromise the target LLM's general capabilities, such as robustness and safety~\cite{qi2024fine}.} Third, some methods~\cite{chen2024magicore, cheng2025think, wang2024make} \textcolor{blue}{directly use or fine-tune} an auxiliary LLM to assess input question difficulty. \textcolor{blue}{However, these approaches do not establish the relationship between the target MLLM's own state and the difficulty of the input question, which limits their ability to accurately assess the input difficulty for the target MLLM.}

In this paper, we examine an intuitive yet underexplored idea, using hidden representations, rather than generated outputs, to estimate question difficulty for LLMs. This idea originates from findings that hidden representations give a finer‑grained and semantically richer view of the model’s prediction logic~\cite{zhang2024reef, yin2024lofit, kong2024aligning}. To conduct an initial exploration of their effectiveness, we extract hidden representations with  Qwen2.5-VL-7B-Instruct~\cite{bai2025qwen2} and InternVL3-8B~\cite{zhu2025internvl3exploringadvancedtraining} for questions in the MMBench dataset~\cite{liu2024mmbench}. As shown in Figure~\ref{fig:problem}, these representations clearly separate easy from hard questions, suggesting that hidden representations may carry information related to question difficulty. 

%=== In this paper,  we address the limitations of existing methods. Achieving this goal requires capturing finer-grained signals from the target MLLM. Rather than relying on generated output of the target MLLM, we focus on the hidden state representations. These representations  provide a more fine-grained and semantically rich perspective~\cite{zhang2024reef, yin2024lofit, kong2024aligning}, reflecting the model’s prediction logic. To explore the relationship between input question difficulty and hidden state representations, we utilize Qwen2.5-VL-7B-Instruct~\cite{bai2025qwen2} and InternVL3-8B~\cite{zhu2025internvl3exploringadvancedtraining} to extract the hidden state representations of the input questions from MMBench dataset~\cite{liu2024mmbench}. As illustrated in Figure~\ref{fig:problem}, the hidden state representations exhibit a clear separation between easy and hard questions. These observations imply that hidden state representations may carry information related to the difficulty of input questions.   

Specifically, we present a method that models input question difficulty by exploiting the hidden representations produced by the target LLM. As the model generates its response token by token, the resulting sequence of hidden representations reflects its reasoning dynamics throughout the generation process. We formalize this sequence of hidden representations as a Markov chain~\cite{norris1998markov}, where the state transitions inherently reflect the model’s autoregressive generation process. To quantify the perceived difficulty at each state, we define a value function over the Markov chain to estimate the expected output quality associated with each state. At test time, the difficulty can be accurately estimated by evaluating the value of the initial state, which is determined solely by the input question, without generating any tokens. 

%=== Building on this insight, we propose a simple yet effective method, named XXX, which models input question difficulty by leveraging the hidden states produced by the target MLLM. As the model generates responses token by token, it produces a sequence of hidden state representations that reflect its evolving internal reasoning.  We formulate this sequence as a Markov chain~\cite{norris1998markov}, and transitions between states capture the model’s autoregressive behavior. \textcolor{red}{To quantify question difficulty, we define a value function over the hidden states, which estimates the expected future output quality based on the current trajectory.} In practice, we train a lightweight two-layer multilayer perceptron (MLP) to learn this value function from the hidden state.

We have conducted extensive experiments to validate the effectiveness of the proposed method across diverse tasks, including general-purpose reasoning, mathematical reasoning and open-ended problem solving, covering both pure textual and multimodal (image-text) scenarios. In addition to accurately estimating input question difficulty for the target LLM, we further validate the effectiveness of our method in adaptive reasoning scenarios. Specifically, we refine repeat-sampling strategies such as Self-Consistency~\cite{wang2022self}, Best-of-N~\cite{brown2024large}, and Self-Refine~\cite{madaan2023self} by adjusting the reasoning strategy based on the estimated difficulty of each question. Experimental results demonstrate that our method provides accurate difficulty estimations, enabling adaptive reasoning to achieve higher inference efficiency across multiple datasets.

%=== In addition to accurately estimating input question difficulty for the target MLLM, we further apply our method to enable adaptive inference in multimodal tasks such as mathematical reasoning, scientific question answering, and open-ended problem solving. Specifically, we focus on improving repeat-sampling-based reasoning strategies, including Self-Consistency~\cite{wang2022self}, Best-of-N~\cite{brown2024large}, and Self-Refine~\cite{madaan2023self}, by dynamically adjusting sampling effort according to the estimated difficulty. [performance]

Overall, our approach enables accurate difficulty estimation without generating multiple outputs at test time, while effectively leveraging fine-grained signals from the model itself and preserving its general capabilities.

%we capture the autoregressive nature of the MLLM inference process and model it as a Markov Decision Process~\cite{williams2007reinforcement} defined over the sequence of hidden states. Based on this formulation, we introduce a recursive difficulty function that quantifies the expected reward at each state transition. During training, we approximate this recursive function by optimizing a value estimator trained on sampled decoding trajectories via a temporal difference learning objective. At inference time, our method directly predicts question difficulty based on the hidden states without the need for explicit token generation, thus enabling efficient difficulty assessment.

%our approach leverages the observed distributional differences in hidden representations as a representation-level signal to support more effective reasoning. During training, we collect the model’s hidden representations and learn a value function that captures their differences based on the model’s output correctness. At inference time, these learned differences are used to guide the adaptive allocation of computational resources, enabling the model to reason more effectively on a per-problem basis while maintaining accuracy and reducing inference cost.

\section{Related Work}

\subsection{Representations of LLMs}

Recent studies have increasingly focused on the hidden representations of LLMs, rather than solely on their textual outputs.. These representations have been explored in various contexts, including alignment~\cite{kong2024aligning, liu2025iterative, li2023inference, zhang2024real}, safety~\cite{chen2025seer, wang2024probing, lu2025x}, interpretability~\cite{ghandehariounpatchscopes, jacobi2025superscopes}, and robustness~\cite{lad2024remarkable, yan2024contrastive}.

\subsection{Repeated Sampling Methods} 

Repeated sampling has been widely employed to enhance the reliability and quality of LLM outputs. Techniques such as Best-of-N~\cite{brown2024large}, Self-Consistency~\cite{wang2022self} and Self-Refine~\cite{madaan2023self} explore multiple generation paths during inference and select or aggregate the results to improve performance.

\subsection{Difficulty Estimation}

A growing body of research explores strategies for adapting model behavior at test time based on input difficulty, with the goal of balancing performance and computational efficiency. These approaches typically rely on categorizing problem difficulty. One prominent line of work focuses on leveraging self-consistency~\cite{wang2022self} in model outputs~\cite{lee2025semantic, pan2024dynathink, li2024escape, chen2024magicore, snell2024scaling}. However, such methods often rely on generating multiple outputs at test time. Alternatively, some methods directly employ or fine-tune an auxiliary large language models as judges to  assess question difficulty~\cite{chen2024magicore, cheng2025think, wang2024make}. However, such auxiliary models often fail to accurately capture the target model’s perception of question difficulty. Another class of approaches fine-tunes LLMs with output-level supervision~\cite{manvi2024adaptive, huang2025efficient}. Although these methods can improve performance on specific reasoning tasks, they compromise the target LLM’s general capabilities.

%  utilizing intrinsic uncertainty to estimate input difficulty without requiring external signals

\clearpage

\section{Methods}

The difficulty of a question, as perceived by the target LLM, is reflected in the quality of its output. One promising approach to accurately estimate perceived difficulty while preserving the general capabilities of the target LLM is to exploit its fine-grained internal signals. This idea offers a potential for model-driven difficulty estimation without requiring multiple outputs at test time. 
% Our methods seek to answer the key question: \textit{How can we accurately estimate perceived difficulty using only the internal signals of the target LLM?}

Estimating perceived difficulty using internal signals requires not only fine-grained modeling of the generation behavior at the token level within the LLM but also sophisticated control over the transitions between generation states across all internal representations. To tackle this challenge, we model the generation process as a Markov chain with token-level hidden representations(Section~\ref{sec:prelim}). We then define a value function to capture the expected output quality at each Markov state (Section~\ref{diff-awre}). Finally, we train a model to learn this value function (Section~\ref{training}), allowing us to estimate the expected output quality based on the initial state determined by the input question.

\subsection{Preliminary}
\label{sec:prelim}

The behavior of transformer-based large language models (LLMs) can be modeled as a Markov chain~\cite{norris1998markov, zekri2024large, kong2024aligning, kao2025safety}. In this paper, we formalize the autoregressive generation process of an LLM as a Markov chain over hidden representations. This formulation is motivated by two key factors. First, hidden representations offer fine-grained internal signals that reflect the model’s evolving reasoning process~\cite{zhang2024reef, yin2024lofit, kong2024aligning}. Second, the Markov chain formulation provides a structured and expressive way to capture the transitions between hidden representations during generation. By modeling these transitions, we can establish a more direct connection between the initial state, which represents the input question, and the eventual output quality.

Formally, for an autoregressive LLM \(f_\theta\) and an input question \(\mathbf{x}\), the generation process at time step \(t\) can be described as:
\[
\begin{aligned}
H_{t+1}, y_{t+1} = f_{\theta}(H_t, y_t),
\end{aligned}
\]
where \(y_t\) is the token generated at time step \(t\). \(H_t = (h_0, h_1, h_2, \ldots, h_t)\) represents the sequence of contextual embeddings up to time \(t\), where each \(h_i\) is the hidden representations generated at time step \(i\) by the LLM. In the formulation, $\{H_t, y_t\}$ serves as the Markov state $s_t$, and \(f_\theta\) acts as the transition function that governs the evolution of the process over time. Specifically, \(s_0 = \{H_0, \mathbf{x}\}\) refers to the initial Markov state, where \(H_0=(h_0)\) is the hidden representations generated from the input question \(\mathbf{x}\).

\subsection{Difficulty Estimation}
\label{diff-awre}

\textbf{To assess perceived difficulty, we need to evaluate the output quality of the target LLM.} This output quality is commonly evaluated using outcome reward models (ORMs), which assess the final result after generation~\cite{coste2023reward, moskovitz2023confronting}. Since that, we define  the reward function \( R \) over the Markov state. Formally, given an input question \( \mathbf{x} \) and an output sequence \( \mathbf{y} = \{y_1, y_2, \dots, y_{t}\} \), we define the reward function \( R \) as:
\begin{equation}
\label{eq:reward}
R(s_t) = 
\begin{cases} 
\text{Reward}(\mathbf{y}) & \text{if } y_t = \text{EOS}, \\
0 & \text{otherwise},
\end{cases}
\end{equation}
where \( \text{EOS} \) denotes the end-of-sequence token.

\textbf{To model the expected output quality from a Markov state, we employ the value function \(V(s_t)\).}  The function models the expected cumulative future rewards that can be obtained by starting from the given state $s_t$. Formally, it can be  defined using the Bellman equation~\cite{bellman1957dynamic} as below:

\begin{equation}
\label{eq:bellman}
V(s_t) = \mathbb{E}_{s_{t+1}}\left[ R(s_t) + \gamma V(s_{t+1}) \right].
\end{equation}

Substituting Eq.~\ref{eq:reward} into Eq.~\ref{eq:bellman}, we obtain:

\begin{equation}
\small
\label{eq:final}
V(s_t) =
\begin{cases}
\gamma \, \mathbb{E}_{s_{t+1}} \left[V(s_{t+1})\right], & \text{if } y_t \neq \text{EOS}, \\
R(s_t), & \text{if } y_t = \text{EOS}.
\end{cases}
\end{equation}

This \( V(s_t) \) represents the expected output quality for the Markov state \( s_t \),  based on the LLM's sample strategy.

\textbf{To estimate the difficulty of an input question \( \mathbf{x} \), we rely solely on the initial state \( s_0 \), which encodes the model's hidden representations for the input question $\mathbf{x}$.} Given the formulation of the value function, the value \( V(s_0) \) represents the expected output quality of the initial Markov state $s_0$ under the LLM's sampling strategy. This directly reflects the perceived difficulty of the question by the target LLM, based solely on the input question itself.

Therefore, the overall difficulty of the problem is defined based on \( V(s_0) \), and the input question \( \mathbf{x} \) is estimated accordingly as:

\[
\mathbf{x} \in 
\begin{cases} 
\text{Difficult} & \text{if } V(s_0) \leq \tau, \\
\text{Easy} & \text{if } V(s_0) > \tau,
\end{cases}
\]
where \( \tau \) is a predefined threshold that separates easy and difficult questions.

\subsection{Training Objective}
\label{training}

\textbf{Our objective is to train a model \( \hat{F}_\phi \) that accurately approximates the value function \( V \).} Given a training dataset \( D = \{\mathbf{x}^{(i)}\}_{i=1}^n \),we first use the target LLM to generate responses. For each input \( \mathbf{x}^{(i)} \), we generate a token sequence \( \mathbf{y}^{(i)} = (y_1^{(i)}, y_2^{(i)}, \ldots, y_t^{(i)}) \).

We interpret our training objective through the lens of temporal difference (TD) learning~\cite{sutton1988learning}. Leveraging the structure in Eq.~\ref{eq:final}, we define the TD error at each step as the discrepancy between the predicted value and the bootstrapped target. Specifically, the TD error is given by:

\[
\delta_t =
\begin{cases}
\gamma \, \text{Reward}(\mathbf{y}^{(i)}) - \hat{F}_\phi(s_t^{(i)}), & \text{if } y_t = \text{EOS}, \\
\gamma \, \hat{F}_\phi(s_{t+1}^{(i)}) - \hat{F}_\phi(s_t^{(i)}), & \text{otherwise}.
\end{cases}
\]

We then minimize the squared TD error over all sampled responses:

\[
\mathcal{L}_{\text{TD}} = \mathbb{E}_{(x, y) \sim \mathcal{D}} \left[ \sum_{t} \delta_t^2 \right].
\]

%\textcolor{blue}{which allows the model to incorporate future information during training through temporal difference updates.}

\section{Application: Difficulty-Aware Repeated Sampling}
\label{sampling}

In this section,  we propose three straightforward strategies to empirically evaluate the performance of our proposed difficulty estimation method in the context of adaptive inference, based on Self-Consistency~\cite{wang2022self} (Section~\ref{self-c}), Best-of-\(N\)~\cite{snell2024scaling} (Section~\ref{BoN}), and Self-Refine~\cite{madaan2023self} (Section~\ref{sr}). Our design principle follows an adaptive strategy: complex repeated sampling methods are employed for questions identified as difficult, whereas easy questions are addressed with single sampling.

\subsection{Difficulty-Aware Self-Consistency}

\label{self-c}
Given question \( x \) and the initial state $s_0$,  we apply a difficulty-aware Self-Consistency strategy. we sample \( K \) reasoning paths 
\[
\{\mathbf{y}^{\text{CoT}}_1, \mathbf{y}^{\text{CoT}}_2, \dots, \mathbf{y}^{\text{CoT}}_K \} 
\sim f_\theta(\mathbf{y} \mid x).
\]

The final prediction is selected adaptively based on the estimated difficulty of \( x \):
\[
\resizebox{\linewidth}{!}{$
\hat{a} =
\begin{cases}
f_\theta^{\text{Direct}}(x), & \text{if } \hat{F}_\phi(s_0) > \tau \\
\arg\max\limits_{a} \sum\limits_{k=1}^K \mathbb{I}(\text{Ans}(\mathbf{y}^{\text{CoT}}_k) = a), & \text{if } \hat{F}_\phi(s_0) \leq \tau
\end{cases}
$}
\]

Here, \( \text{Ans}(\cdot) \) extracts the final answer from a chain-of-thought (CoT) reasoning path.

\subsection{Difficulty-Aware Best-of-$N$}

\label{BoN}
Given question \( x \) and its initial state  $s_0$, we apply a difficulty-aware Best-of-\(N\) strategy. we sample \( K \) reasoning paths 
\[
\{\mathbf{y}^{\text{CoT}}_1, \mathbf{y}^{\text{CoT}}_2, \dots, \mathbf{y}^{\text{CoT}}_K \} 
\sim f_\theta(\mathbf{y} \mid x).
\]

and then select the one with the highest predicted correctness score:
\[
\resizebox{\linewidth}{!}{$
\hat{a} =
\begin{cases}
f_\theta^{\text{Direct}}(x), & \text{if } \hat{F}_\phi(s_0) > \tau \\
\arg\max\limits_{k = 1,\dots,K} \; P(\text{true} \mid x, \mathbf{y}_k^{\text{CoT}}), & \text{if } \hat{F}_\phi(s_0) \leq \tau
\end{cases}
$}
\]

Here, \( P(\text{true} \mid x, y_k) \) denotes the model’s estimated probability that candidate \( y_k \) is a correct answer to input \( x \). 

\subsection{Difficulty-Aware Self-Refine}

\label{sr}
Given question \( x \) and its initial state  $s_0$, we apply a difficulty-aware Self-Refine strategy. We first sample an initial response \( y_0 \sim f_\theta(y \mid x) \), and then iteratively refine it for \( T \) steps to obtain a sequence of improved responses:
\[
\resizebox{\linewidth}{!}{$
\{\mathbf{y}_1^{\text{CoT}}, \mathbf{y}_2^{\text{CoT}}, \dots, \mathbf{y}_t^{\text{CoT}}\}, 
\quad \text{where} \quad 
\mathbf{y}_t \sim f_\theta(\mathbf{y} \mid x, \mathbf{y}_{t-1}).
$}
\]

The final prediction is chosen adaptively based on the estimated difficulty of \( x \):
\[
\hat{a} =
\begin{cases}
f_\theta^{\text{Direct}}(x), & \text{if } \hat{F}_\phi(s_0) > \tau \\
\text{Ans}(\mathbf{y}^{(T)}), & \text{if } \hat{F}_\phi(s_0) \leq \tau
\end{cases}
\]
Here, \( \text{Ans}(\cdot) \) extracts the final answer from the last refined reasoning path.

\clearpage
\section{Experiments}

In this section, we conduct experiments to evaluate our method. Our experiments are divided into two main parts. The first part assesses whether our proposed method can accurately estimate the difficulty of input questions as perceived by the target LLMs (Section~\ref{sec:result-diffestimate}). The second part examines whether applying our difficulty estimation to difficulty-aware repeated sampling improves performance (Section~\ref{sec:result-diffapplication}).

\subsection{Experiment Setup}

\label{exp:setup}

\noindent\textbf{Datasets.} We select six datasets to evaluate the proposed method, covering both multimodal (image-text) and purely textual settings, and encompassing both general-purpose and domain-specific question answering tasks. For general-purpose reasoning QA, we use MMBench~\cite{MMBench}, ScienceQA~\cite{lu2022learn}, commonsenseQA~\cite{talmor2018commonsenseqa} and strategyQA~\cite{geva2021did}. For domain-specific reasoning in mathematics QA, we use MathVista~\cite{lu2024mathvista} and gsm8k~\cite{cobbe2021training}. More details can be found in Appendix~\ref{dataset-details}.

\noindent\textbf{Baselines.} We compare our method with below methods:

\begin{itemize}
    \item prompt: A method that estimates problem difficulty by using a prompt to instruct the model to assess the difficulty of the input question itself.
    \item AG~\cite{lee2025semantic}: A method that estimates problem difficulty based on the consistency of the target model’s outputs.
    \item LLMs-Ranking~\cite{wang2024make}: A method that introduces an auxiliary LLM to directly assess the difficulty of a given problem.
    \item HaluSearch-Gen~\cite{cheng2025think}: A training-based method that fine-tunes the LLM to equip it with the ability to assess problem difficulty.
    \item HaluSearch-Critic~\cite{cheng2025think}: A variant of HaluSearch-Gen that incorporates a critic signal into the training data to provide more explicit supervision for difficulty assessment.

We provide more details about baseline in the Appendix~\ref{baseline-details}.

\end{itemize}

\noindent\textbf{Models.} To evaluate the effectiveness of our proposed method on both text-only and multimodal datasets, we employ two advanced open-source multimodal large language models in our experiments: Qwen2.5-VL-7B-Instruct~\cite{bai2025qwen2} and InternVL3-8B~\cite{zhu2025internvl3exploringadvancedtraining}.

\noindent\textbf{Evaluation Metrics.} (1) \textbf{Difficulty Estimation.} To evaluate the accuracy of our method in the binary classification task of question difficulty estimation, we use ROC-AUC~\cite{gonen2006receiver} and Macro-F1 score~\cite{sokolova2009systematic}. These two metrics provide a comprehensive assessment of classification accuracy, even in the presence of severe class imbalance. (2) \textbf{Difficulty-Aware Repeated Sampling.} We use accuracy and total output length to evaluate the efficiency of the proposed difficulty estimation method when applied to the difficulty-aware repeated sampling. Accuracy measures the proportion of correctly answered questions, reflecting the overall performance, while total output length reflects the cost of the sampling process.

\noindent\textbf{Implementation Details.} (1) \textbf {Ground Truth.} Since the datasets are all objective questions, we use the model's responses in three independent attempts to determine whether it answers all questions correctly, which serves as the ground truth for classifying the input questions as easy or hard.  (2) \textbf{Model Training.} We employ a two-layer fully connected neural network to implement the difficulty estimation described in Section~\ref{training}. In practice, we use the last layer hidden state \( h_t \) as the state \( s_t \) at time step $t$. More details about implementation are shown in Appendix~\ref{imply-details}.

\begin{table*}[t]
\centering
\resizebox{\textwidth}{!}{
\begin{tabular}{cccccccccc}
\toprule
                            &                                        & \multicolumn{4}{c}{\textbf{Qwen2.5vl-7B-Instruct}}                                  & \multicolumn{4}{c}{\textbf{InternVL3-8B}}          \\ \cline{3-10} 
\textbf{Dataset}                    & \textbf{Method}                                & \textbf{Easy-Acc} & \textbf{Hard-Acc} & \textbf{ROC-AUC}        & \textbf{Macro-F1}           & \textbf{Easy-Acc} & \textbf{Hard-Acc} & \textbf{ROC-AUC} & \textbf{Macro-F1} \\ \midrule
\multirow{6}{*}{MMBench}    & \multicolumn{1}{c|}{prompt}            &   89.90      &     6.48      & 51.75   & \multicolumn{1}{c|}{47.75}               &      40.94    &   37.39    &   60.03     &    33.61      \\
                            & \multicolumn{1}{c|}{AG}                & 88.27    & 61.90    & 81.04              & \multicolumn{1}{c|}{72.17}          & 98.59    & 11.76    & 55.99   & 56.62    \\
                            & \multicolumn{1}{c|}{LLMs-Ranking}      & 74.25    & 33.45    & 44.28          & \multicolumn{1}{c|}{51.43}          &     98.59     &    2.10      &    42.49     &    48.65     \\
                            & \multicolumn{1}{c|}{HaluSearch-Gen}    & 37.95    & 86.23    & 37.39          & \multicolumn{1}{c|}{41.23}          &     91.96     &   39.07      &  70.89       &    65.23      \\
                            & \multicolumn{1}{c|}{HaluSearch-Critic} & 69.41    & 51.08    & 38.99          & \multicolumn{1}{c|}{53.48}          &      88.27    &   57.14       &   76.37   &   68.27    \\
                            & \multicolumn{1}{c|}{\textbf{ours}}     & 91.08    & 78.81    & \textbf{94.15} & \multicolumn{1}{c|}{\textbf{80.68}} &     88.35     &   73.93    &      \textbf{91.22}   &     \textbf{75.98}     \\ \midrule
\multirow{6}{*}{ScienceQA}  & \multicolumn{1}{c|}{prompt}            &   98.51     &     1.00       &          50.24      & \multicolumn{1}{c|}{46.14}               &     56.81 &   32.09    &    55.54   &  39.85    \\
                            & \multicolumn{1}{c|}{AG}                & 86.29    & 59.82    & 77.76              & \multicolumn{1}{c|}{70.40}          &     97.64     &    20.10      &  59.07       &    61.47      \\
                            & \multicolumn{1}{c|}{LLMs-Ranking}      & 56.47    & 55.33    & 42.06          & \multicolumn{1}{c|}{48.83}          &       99.34   &   1.87  &  37.72       &    49.60      \\
                            & \multicolumn{1}{c|}{HaluSearch-Gen}    & 93.64    & 18.06    & 44.27          & \multicolumn{1}{c|}{56.68}          &    96.30      &   39.87       &   70.50  &  69.41        \\
                            & \multicolumn{1}{c|}{HaluSearch-Critic} & 66.37    & 49.55    & 41.49          & \multicolumn{1}{c|}{52.82}          &     93.09     &    61.68      &   79.44      &   72.50       \\
                            & \multicolumn{1}{c|}{\textbf{ours}}     & 90.00       & 76.15    & \textbf{93.09} & \multicolumn{1}{c|}{\textbf{79.48}} & 89.61         &   73.60       &     \textbf{92.02}    &   \textbf{74.62}       \\ \midrule
\multirow{6}{*}{MathVista}  & \multicolumn{1}{c|}{prompt}            &   76.90       &   10.76       &       51.32         & \multicolumn{1}{c|}{41.95}               &    40.94   &   58.57   &   51.79   &   46.94   \\
                            & \multicolumn{1}{c|}{AG}                & 65.79    & 76.58    & 78.71              & \multicolumn{1}{c|}{67.81}          & 92.22    & 24.29    & 59.79   & 58.44    \\
                            & \multicolumn{1}{c|}{LLMs-Ranking}      & 0.00        & 100.00      & 42.91          & \multicolumn{1}{c|}{25.71}          & 93.61    & 7.86     &    47.81     &  47.12        \\
                            & \multicolumn{1}{c|}{HaluSearch-Gen}    & 100.00      & 0.60      & 49.66          & \multicolumn{1}{c|}{42.25}          &   46.39      &  65.00    &   53.42     &  50.46    \\
                            & \multicolumn{1}{c|}{HaluSearch-Critic} & 69.66    & 45.07    & 41.71          & \multicolumn{1}{c|}{51.53}          &     66.39     &   47.86      &    57.98      &   55.99        \\
                            & \multicolumn{1}{c|}{\textbf{ours}}     & 85.63    & 64.83    & \textbf{86.13} & \multicolumn{1}{c|}{\textbf{75.23}} &        84.21  &    71.52   &     \textbf{86.11}    & \textbf{77.44}         \\ \midrule
\multirow{6}{*}{StrategyQA}     & \multicolumn{1}{c|}{prompt}            &     97.39     &   0.38    &      51.12     & \multicolumn{1}{c|}{37.77}          &    59.02   &  27.73      &     54.03    & 43.43         \\
& \multicolumn{1}{c|}{AG}            &     73.16     &  34.96    &      56.95     & \multicolumn{1}{c|}{53.85}          &    92.65   &  9.66      &     51.16    & 46.34         \\
                            & \multicolumn{1}{c|}{LLMs-Ranking}      & 0.00    &  100.00   & 55.35          & \multicolumn{1}{c|}{27.91}          &   88.42       &  17.23   &    58.83     &  50.45       \\
                            & \multicolumn{1}{c|}{HaluSearch-Gen}    & 84.89    & 27.41    & 56.87         & \multicolumn{1}{c|}{54.78}       &  83.86 & 29.41  & 60.04 & 55.53     \\
                            & \multicolumn{1}{c|}{HaluSearch-Critic} &      69.78   &     48.51     &  61.58  & \multicolumn{1}{c|}{59.21}     &   78.07  & 37.50  & 61.06 & 57.58      \\
                            & \multicolumn{1}{c|}{\textbf{ours}}     & 58.99    & 75.18    & \textbf{73.09} & \multicolumn{1}{c|}{\textbf{65.22}} &    68.13      & 62.50    &   \textbf{70.95} &     \textbf{64.16}     \\ \midrule
\multirow{6}{*}{gsm8k} & \multicolumn{1}{c|}{prompt}            &  99.68  &     0.52     &      49.89        & \multicolumn{1}{c|}{42.01}               &   67.69       &     59.09    &     38.38    &  56.93        \\
& \multicolumn{1}{c|}{AG}            &  51.54  &     68.58     &      65.60        & \multicolumn{1}{c|}{55.23}               &   92.36       &   13.18    &     52.77    &  52.74        \\
                            & \multicolumn{1}{c|}{LLMs-Ranking}      & 0.00    & 100.00    &   51.21    & \multicolumn{1}{c|}{22.46}          &   97.99       &   4.55       &   67.57      &    49.11      \\
                            & \multicolumn{1}{c|}{HaluSearch-Gen}    & 91.10    & 24.17   & 58.91          & \multicolumn{1}{c|}{55.70}           &      65.72    &  43.12      &  59.21    & 53.67         \\
                            & \multicolumn{1}{c|}{HaluSearch-Critic} &   98.01     &  18.68      &  62.21  &   \multicolumn{1}{c|}{57.87}             &   84.23  &  34.21 &  58.97   &    54.43      \\
                            & \multicolumn{1}{c|}{\textbf{ours}}      &  61.67       &  58.24 &    \textbf{67.28}  & \multicolumn{1}{c|}{\textbf{65.23}} &   63.02       &    63.10      &     \textbf{67.63}    &  \textbf{61.90}        \\ \midrule
\multirow{6}{*}{commonsenseQA} & \multicolumn{1}{c|}{prompt}            &  97.59  &     2.26    &      45.44        & \multicolumn{1}{c|}{50.08}               &   62.74       &     36.67    &     50.79    &  46.37        \\
& \multicolumn{1}{c|}{AG}            &  78.43  &    62.78     &      73.72        & \multicolumn{1}{c|}{67.68}               &   96.04       &   13.51    &     54.78    &  55.37        \\
                            & \multicolumn{1}{c|}{LLMs-Ranking}      & 0.00    & 100.00    & 53.77      & \multicolumn{1}{c|}{17.89}          &   97.20       &   1.62       &   46.64      &    50.05      \\
                            & \multicolumn{1}{c|}{HaluSearch-Gen}    & 86.02    & 19.86   & 53.08          & \multicolumn{1}{c|}{52.91}           &   91.06       &   16.36     &  54.23   &  53.93        \\
                            & \multicolumn{1}{c|}{HaluSearch-Critic} &   58.88     &  46.26      &  52.46  &   \multicolumn{1}{c|}{48.19}             &         71.10 & 58.69  &  53.71   &  51.72        \\
                            & \multicolumn{1}{c|}{\textbf{ours}}      &  72.35     &  73.64 &    \textbf{80.78}  & \multicolumn{1}{c|}{\textbf{69.81}} &  76.46        &      74.30    &   \textbf{83.00}      &  \textbf{68.10}        \\ \bottomrule
\end{tabular}
}
\caption{Performance comparison between our method and other approaches. \textquotedblleft Easy-Acc\textquotedblright refers to the proportion of questions classified as easy out of all the easy questions, while \textquotedblleft Hard-Acc\textquotedblright refers to the proportion of questions classified as easy out of all the hard questions.}
\label{tab:performance-difficulty}
\end{table*}

\subsection{Main Results for Difficulty Estimation}
\label{sec:result-diffestimate}
To assess whether our method provides an accurate estimation of input question difficulty, we conduct experiments on six diverse datasets using two state-of-the-art models.

\textbf{Our method provides accurate difficulty estimation.} From the results presented in Table~\ref{tab:performance-difficulty}, it is clear that our method nearly outperforms the baseline methods across all evaluated metrics. In terms of difficulty estimation, our method achieves strong performance in both easy and hard question identification. Our method excels in providing precise difficulty classification, achieving high ROC-AUC and Macro-F1 scores. For instance, on the ScienceQA dataset, our method reaches an ROC-AUC of 93.09\% and a Macro-F1 of 79.48\%, demonstrating its excellent ability to classify both easy and hard questions accurately. 

Among the baselines, AG yields the strongest performance in most cases. This is primarily due to its consistency-based evaluation strategy, which leverages the output information of the target LLM itself. In addition, LLM-Ranking exhibits less stable performance, which can be attributed to two factors. First, it relies on an auxiliary model to assess the difficulty directly from the input, without leveraging the internal characteristics of the target LLM. Our method, by leveraging the internal hidden representations of the target LLM, offers a more direct and fine-grained reflection of its reasoning process, resulting in improved stability and generalization across datasets.

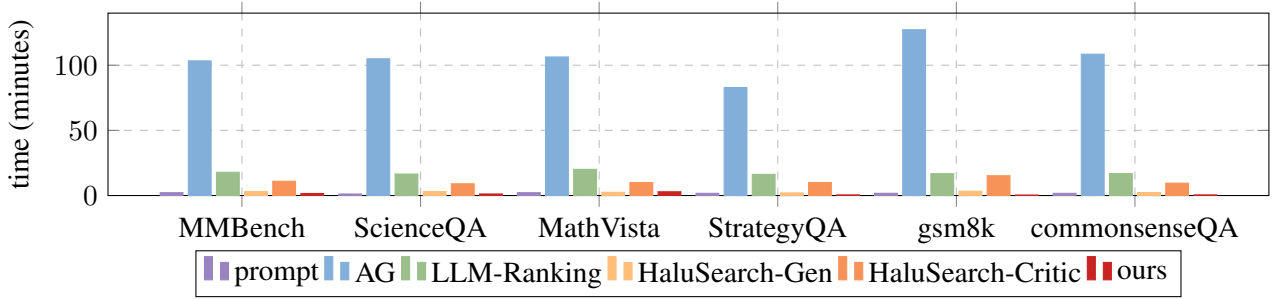
\begin{figure*}
    \centering
    \begin{tikzpicture}
    \begin{axis}
       [ybar , %柱状图 xbar 条形图
    grid=major,major grid style={dashed}, %显示背景网格
    ymin=0,  %Y轴刻度最小值
    ylabel=time (minutes),  %Y轴标签
    bar width=.3cm, %柱子宽度
    width=60cm,
    height=4cm,  %图片的长和宽
    symbolic x coords={MMBench, ScienceQA, MathVista, StrategyQA, gsm8k,commonsenseQA},  %将x轴刻度设置为指定符号
    x=2.35cm, %设置x轴每个单元格的长度
    xtick=data, %根据数据设置x轴刻度
    nodes near coords style={font=\fontsize{8}{12}\selectfont}, %数字大小为8pt，行距为12pt
    enlarge x limits=0.15, %将x轴范围增加20%
    legend style={at={(0.5,-0.3)},anchor=north,legend columns=-1}, %将标签放在{(0.5,-0.2)}表示图例的中心，anchor=north表示图例的顶部与底部对齐，legend columns=-1表示图例横排
    ] 
    %添加数据 
    \addplot+ [c1, draw=c1]coordinates {(MMBench, 2.19) (ScienceQA, 1.18) (MathVista, 2.195) (StrategyQA, 1.70) (gsm8k, 1.72) (commonsenseQA, 1.69)};
    \addplot+ [c2, draw=c2]coordinates {(MMBench, 103.36) (ScienceQA, 104.96) (MathVista, 106.35) (StrategyQA, 82.9015) (gsm8k, 127.27) (commonsenseQA, 108.5)};
    \addplot+ [c3, draw=c3]coordinates {(MMBench, 17.87) (ScienceQA, 16.50) (MathVista, 20.11) (StrategyQA, 16.26) (gsm8k, 16.78) (commonsenseQA, 16.83)}; 
    \addplot+ [c4, draw=c4]coordinates {(MMBench, 3) (ScienceQA, 3) (MathVista, 2.5) (StrategyQA, 2) (gsm8k, 3.3) (commonsenseQA, 2.3)};
    \addplot+ [c5, draw=c5]coordinates {(MMBench, 11) (ScienceQA, 9) (MathVista, 10) (StrategyQA, 10) (gsm8k, 15.2) (commonsenseQA, 9.5)};
     \addplot+ [c6, draw=c6]coordinates {(MMBench, 1.57) (ScienceQA, 1.2) (MathVista, 2.89) (StrategyQA, 0.5) (gsm8k, 0.4) (commonsenseQA, 0.5)};
    \legend{prompt, AG, LLM-Ranking, HaluSearch-Gen, HaluSearch-Critic, ours};
    \end{axis} 
    \end{tikzpicture}
    \caption{Time comparison for difficulty estimation across different datasets and methods at test time for qwen2.5vl-7B-Instruct. To ensure a fair comparison, we randomly selected $400$ questions from the test set of each dataset to evaluate the time cost.}
    \label{fig:estimate_cost}
\end{figure*}

\begin{figure*}[!t]
    \small
    \centering
    \includegraphics[width=0.95\textwidth,height=0.35\textheight]{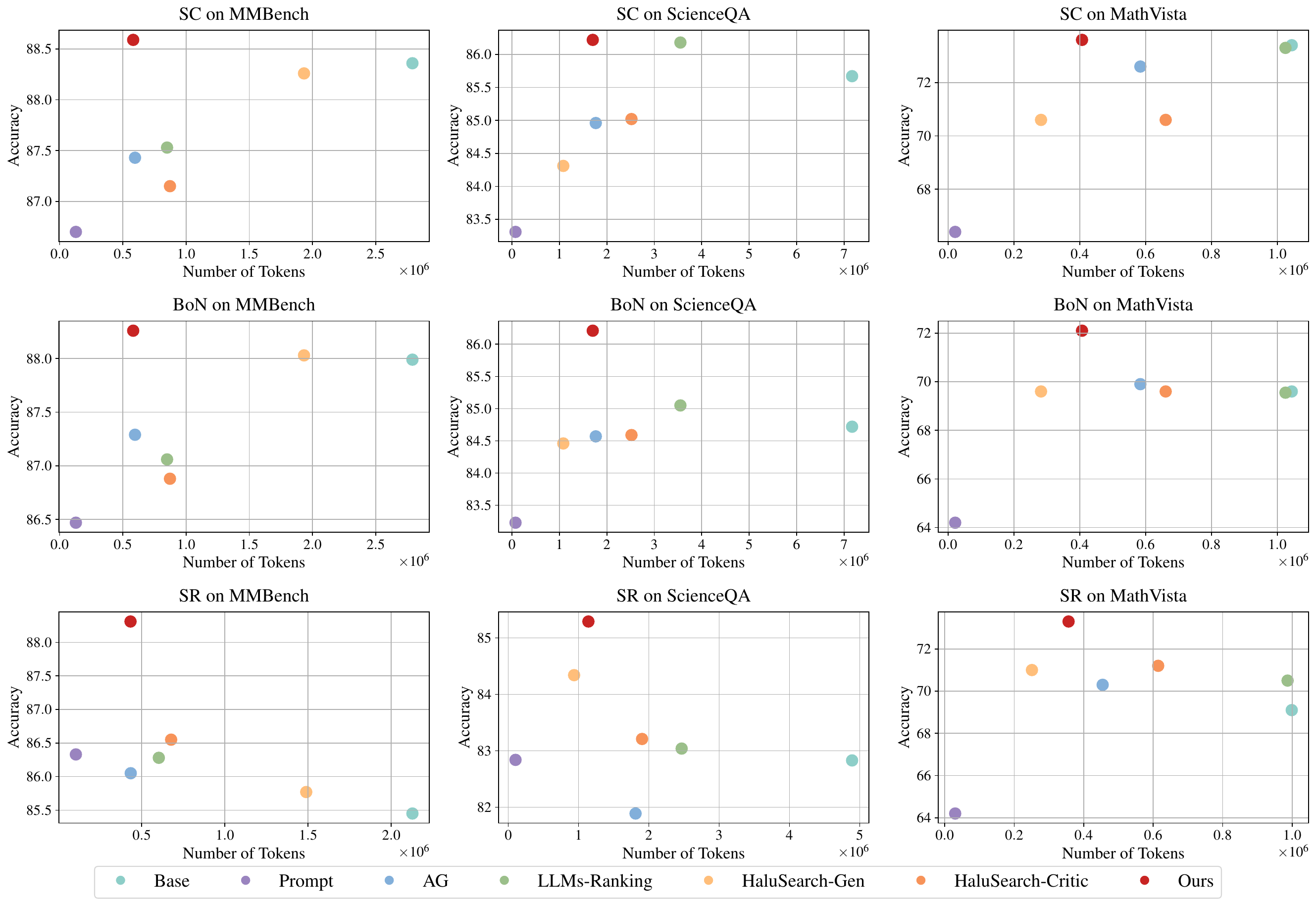}
    \caption{Performance comparison of difficulty-aware sampling methods across multiple datasets for Qwen2.5vl-7B-instruct. "SC" refers to Self-Consistency, "BoN" refers to the Best-of-N, and "SR" refers to Self-Refine. "Number of Tokens" represents the total number of output tokens generated on the test set. For SC and BoN, the budget $K$ is set to $10$, while for SR, the budget $K$ is set to $5$.}
    \label{fig:MMBench}
\end{figure*}

\textbf{The time cost of our method for assessing input question difficulty at test time is low.} This time cost directly affects the efficiency of subsequent tasks that rely on difficulty estimation. Figure~\ref{fig:estimate_cost} presents the evaluation results. As shown, our method consistently requires less time compared to other baselines across all datasets, ensuring efficient difficulty estimation. For prompt, AG, and LLM-Rankings, this training-free method inevitably involves extensive preprocessing of the questions during the test time, resulting in significant time costs. For HaluSearch-Gen and HaluSearch-Critic, although they learn difficulty estimation during training, they still require a single evaluation to allow the model to perceive the difficulty. In contrast, our method only requires the hidden representations generated by the target LLM from the input questions at test time, enabling fast and efficient difficulty assessment.

\textbf{Since our method does not involve training the target LLM itself, it does not impair the model’s general capabilities. }

\begin{table*}[t]
\centering
\resizebox{\textwidth}{!}{
\begin{tabular}{cccccccccc}
\toprule
                            &                                        & \multicolumn{4}{c}{\textbf{Qwen2.5vl-7B-Instruct}}                                  & \multicolumn{4}{c}{\textbf{InternVL3-8B}}          \\ \cline{3-10} 
\textbf{Dataset}                    & \textbf{Method}                                & \textbf{Easy-Acc} & \textbf{Hard-Acc} & \textbf{ROC-AUC}        & \textbf{Macro-F1}           & \textbf{Easy-Acc} & \textbf{Hard-Acc} & \textbf{ROC-AUC} & \textbf{Macro-F1} \\ \midrule
\multirow{5}{*}{RLHF-V}     & \multicolumn{1}{c|}{prompt}            &     99.80     &   0.00      &      50.10          & \multicolumn{1}{c|}{49.14}               &       45.80   &    56.00      &     50.82    & 34.20         \\
                            & \multicolumn{1}{c|}{LLMs-Ranking}      & 60.08    & 36.36    & 47.36          & \multicolumn{1}{c|}{39.49}          &   75.66       &   18.99   &    44.75     &   46.82       \\
                            & \multicolumn{1}{c|}{HaluSearch-Gen}    & 97.17    & 46.15    & 45.83          & \multicolumn{1}{c|}{58.71}          &  94.61  & 34.14  & 82.48 & 66.72     \\
                            & \multicolumn{1}{c|}{HaluSearch-Critic} &       95.09   &     65.85     &  67.32  & \multicolumn{1}{c|}{\textbf{71.94}}     &   94.96  & 65.85  & 67.32 & 75.22      \\
                            & \multicolumn{1}{c|}{\textbf{ours}}     & 79.64    & 70.79    & \textbf{82.84} & \multicolumn{1}{c|}{67.39} &        81.05  &    71.83      &    \textbf{85.95}     &   \textbf{79.83}       \\ \midrule
\multirow{5}{*}{VLFeedback} & \multicolumn{1}{c|}{prompt}            &  90.49  &     11.22     &      49.12        & \multicolumn{1}{c|}{48.43}               &    76.59      &      43.18    &     51.47    &     50.04     \\
                            & \multicolumn{1}{c|}{LLMs-Ranking}      & 85.03    & 16.33    & 49.92          & \multicolumn{1}{c|}{49.91}          &    69.96      &    11.36      &    33.23     &     41.73      \\
                            & \multicolumn{1}{c|}{HaluSearch-Gen}    & 98.99    & 48.78    & 69.36          & \multicolumn{1}{c|}{58.16}           &   96.47      &     6.68   &  50.19    &    55.73      \\
                            & \multicolumn{1}{c|}{HaluSearch-Critic} &    99.07     &  48.61      &    72.72  &   \multicolumn{1}{c|}{62.77}             &    96.04     &       66.66   &     71.84    &    65.79       \\
                            & \multicolumn{1}{c|}{\textbf{ours}}     & 92.39    & 80.50    & \textbf{92.16} & \multicolumn{1}{c|}{\textbf{75.57}} &  93.96        &   73.21       &     \textbf{74.94}   & \textbf{77.82}         \\             
                        \bottomrule
\end{tabular}
}
\caption{Performance comparison between our method and other approaches. }
\label{tab:performance-difficulty-subject}
\end{table*}

\subsection{Main Results for Difficulty-Aware Repeated Sampling}
\label{sec:result-diffapplication}

Figure~\ref{fig:MMBench} shows the performance of different difficulty-aware sampling methods on MMBench, ScienceQA, and MathVista under three sampling strategies: Self-Consistency (SC), Best-of-N (BoN), and Self-Refine (SR). We apply the six baseline methods introduced in the section~\ref{exp:setup} solely for difficulty estimation, and incorporate these difficulty estimation into the difficulty-aware sampling framework proposed in section~\ref{sampling}. 

\textbf{Across all datasets and sampling strategies, our method consistently achieves the highest accuracy while consuming fewer or comparable numbers of tokens.} For example, on ScienceQA under SC, our method reaches the highest accuracy while using fewer tokens than most baselines. On MathVista, our method demonstrates substantial improvements under both BoN and SR, outperforming all other approaches in terms of accuracy. Although the prompt method tends to consume fewer tokens, this is largely due to its inclination to classify most questions as easy. However, such over-simplification often leads to lower overall accuracy, as it fails to allocate sufficient reasoning effort for genuinely difficult questions. 

% \textcolor{red}{Compared with token-intensive baselines such as AG and LLMs-Ranking, our method provides a more favorable trade-off between accuracy and computational cost. These results demonstrate that the proposed difficulty estimation can effectively guide sampling strategies, enabling more accurate and efficient generation.}

The results for other budget are shown in Figure~\ref{fig:MMBench2} in Appendix.

\section{Further Analysis}

In this section, we conduct further analyses to validate the robustness, generalizability, and scalability of our method. 

\subsection{Performance on Open-ended Questions}

To comprehensively assess the applicability of our method beyond open-ended tasks, we evaluate its performance on two open-ended datasets, including RLHF-V~\cite{yu2023rlhf} and VLFeedback~\cite{2023vlfeedback}. We adopt LLaVA-Critic~\cite{xiong2024llava} as the reward model to score the responses generated by the target model. These scores are used as ground-truth difficulty annotations. Specifically, a score below $50$ indicates a hard question, while a score above $50$ denotes an easy one, with $100$ being the highest possible score. The experimental results are presented in Table~\ref{tab:performance-difficulty-subject}. These results demonstrate that our method also performs well in open-ended scenarios.

\begin{table}[]
\resizebox{0.5\textwidth}{!}{
\begin{tabular}{ccccc}
\toprule
Method            & train dataset               & test dataset                                        & ROC-AUC & Macro-F1 \\ \midrule
HaluSearch-Gen    & \multirow{3}{*}{ScienceQA}  & \multicolumn{1}{c|}{\multirow{3}{*}{MMBench}}       &  50.89       &     49.93     \\
HaluSearch-Critic &                         & \multicolumn{1}{c|}{}                               &  58.27       &    46.57      \\
ours              &                             & \multicolumn{1}{c|}{}                            &   \textbf{89.03}      &     \textbf{73.73}     \\ \midrule
HaluSearch-Gen    & \multirow{3}{*}{StrategyQA} & \multicolumn{1}{c|}{\multirow{3}{*}{commonsenseQA}} &      52.16   &     50.98     \\
HaluSearch-Critic &                             & \multicolumn{1}{c|}{}                               &  49.66       &   47.55       \\
ours              &                             & \multicolumn{1}{c|}{}                        &    \textbf{65.69}     &     \textbf{56.08}     \\ \bottomrule
\end{tabular}
}
\caption{Generalization comparison between training-based approaches.}
\label{tab:cross-domain}
\end{table}

\begin{table}[]
\centering
\resizebox{0.5\textwidth}{!}{
\begin{tabular}{ccccc}
\toprule
\multicolumn{1}{l}{}                         & \multicolumn{2}{c}{ScienceQA} & \multicolumn{2}{c}{commonsenseQA}                          \\ \cline{2-5} 
Model                                        & ROC-AUC       & Macro-F1      & \multicolumn{1}{c}{ROC-AUC} & \multicolumn{1}{c}{Macro-F1} \\ \midrule
\multicolumn{1}{c|}{Qwen2.5-VL-3B-Instruct}  &       90.05       &        77.60       &     79.48         &     66.04                         \\
\multicolumn{1}{c|}{Qwen2.5-VL-7B-Instruct}  &     93.04          &   \textbf{79.48}            &         80.78                    &      67.66                    \\
\multicolumn{1}{c|}{Qwen2.5-VL-32B-Instruct} &     94.32          &       79.46         &                 84.02             &   \textbf{70.84}                           \\
\multicolumn{1}{c|}{Qwen2.5-VL-72B-Instruct} &      \textbf{95.40}         &    78.69           &                  \textbf{85.51}           &    70.10                          \\ \bottomrule
\end{tabular}
}
\caption{Difficulty estimation performance across model sizes.}
\label{tab:scalability}
\end{table}

\subsection{Cross-Domain Generalization}

\myparagraph{Generalization Across Datasets}To evaluate the cross-domain generalization capability of our method, we train it on one dataset and directly evaluate it on a different domain. This setting tests whether the learned difficulty estimation can generalize to new types of inputs. We compare our method with a training-based baseline that is trained and evaluated on the same domain. The results are shown in Table~\ref{tab:cross-domain}, demonstrating that our method maintains strong performance even under domain shifts.

\myparagraph{Generalization Across Models}  Beyond dataset-level generalization, we further investigate whether the learned value function can generalize across different LLMs. Generalizing across models is inherently challenging, as different LLMs encode distinct and model-specific feature representations depending on their architecture and training data~\cite{zhang2024reef,dang2024explainable}.  We train on Qwen2.5-7B-Instruct and evaluate on LLaMA-3.1-8B-Instruct.  As shown in Table~\ref{tab:model-generalization}, we observe a notable drop in ROC-AUC and Macro-F1 scores, suggesting that direct transfer across heterogeneous models is limited. To address this, we adopt a strategy of training a separate lightweight value function for each target model. We further conduct experiments to compare the training-time cost of our method with existing approaches. As shown in Table~\ref{tab:training-time}, our method requires significantly less training time compared to  training-based approaches.

\begin{table}[h]
\centering
\resizebox{0.5\textwidth}{!}{
\begin{tabular}{cccccc}
\toprule
Dataset & Trained Model & Test Model & ROC-AUC & Macro-F1 \\ \midrule
commonsenseQA & Qwen2.5-7B-Instruct & LLaMA-3.1-8B-Instruct & 47.29 & 43.21 \\
StrategyQA    & Qwen2.5-7B-Instruct & LLaMA-3.1-8B-Instruct & 49.74 & 40.15 \\ \bottomrule
\end{tabular}
}
\caption{Generalization performance of the value function across different LLMs.}
\label{tab:model-generalization}
\end{table}

\begin{table}[h]
\centering
\resizebox{0.5\textwidth}{!}{
\begin{tabular}{lcc}
\toprule
Method              & MMBench (minutes) & ScienceQA (minutes) \\ \midrule
HaluSearch-Gen      & 5.32              & 5.20 \\
HaluSearch-Critic   & 10.23             & 9.15 \\
Ours                & 1.56              & 1.85 \\ \bottomrule
\end{tabular}
}
\caption{Training-time comparison with baseline methods. $400$ samples were randomly selected for evaluation.}
\label{tab:training-time}
\end{table}

\subsection{Scalability Across Model Sizes}

To evaluate the scalability of our method with respect to model size, we conduct experiments on LLMs of different scales. Specifically, we apply our difficulty estimation framework to models of varying sizes within the same architecture family, including Qwen2.5-VL-3B-Instruct, Qwen2.5-VL-7B-Instruct, Qwen2.5-VL-32B-Instruct, and Qwen2.5-VL-72B-Instruct. All experiments are performed on the ScienceQA dataset to ensure a consistent evaluation setting. The results, presented in Table~\ref{tab:scalability}, show that our method maintains consistently high performance across different model sizes. 

\begin{table}[h]
\centering
\resizebox{0.45\textwidth}{!}{
\begin{tabular}{ccccc}
\toprule
Model & 3 & 4 & 5 & 6 \\ \midrule
Qwen2.5-VL-7B-Instruct & 80.68 & 80.82 & 80.51 & 80.92 \\
InternVL3-8B           & 75.98 & 75.52 & 76.42 & 76.55 \\ \bottomrule
\end{tabular}
}
\caption{Effect of the number of inference attempts on Macro-F1 scores (MMBench).}
\label{tab:sampling-ablation}
\end{table}

\subsection{Ablation Study}

\myparagraph{Ablation on Inference Attempts}We further perform ablation studies to investigate the impact of the number of independent inference attempts in our framework. Specifically, we vary the number of attempts from $3$ to $6$ and evaluate the Macro-F1 scores on the MMBench dataset. As shown in Table~\ref{tab:sampling-ablation}, the results demonstrate that three attempts are already sufficient to capture the variability of model outputs, while additional attempts yield only marginal improvements.

\myparagraph{Ablation on Order of the Markov Process}We also examine the impact of considering higher-order Markov processes by redefining the state as a tuple of the previous $k$ states. Experiments are conducted on the MMBench dataset using Qwen2.5-VL-7B-Instruct. As shown in Table~\ref{tab:markov-ablation}, the differences in ROC-AUC and Macro-F1 are marginal (within $\pm 1$ point), 
indicating that the choice of $k$ has minimal impact on overall performance.

\begin{table}[h]
\small
\centering
\begin{tabular}{ccc}
\toprule
$k$ & ROC-AUC & Macro-F1 \\ \midrule
1 & 94.15 & 80.68 \\
2 & 93.19 & 81.11 \\
3 & 94.07 & 80.76 \\ \bottomrule
\end{tabular}
\caption{Effect of Markov order $k$ on difficulty estimation performance (MMBench).}
\label{tab:markov-ablation}
\end{table}

% \usepackage{multirow}

\myparagraph{Ablation on Answer Extraction Method} 
we also examine the reliability of the answer extraction process. In particular, we replace our original extractor with GPT-4o on the MMBench dataset. As shown in Table~\ref{tab:extraction-reliability}, the results remain highly consistent, 
with differences in ROC-AUC and Macro-F1 within $\pm 1.5$ points. 
This indicates that potential extraction errors caused by formatting inconsistencies or partial reasoning 
have minimal impact on reward labeling and difficulty estimation.

\begin{table}[h]
\centering
\resizebox{0.45\textwidth}{!}{
\begin{tabular}{lcccc}
\toprule
Model & Method & ROC-AUC & Macro-F1 \\ \midrule
Qwen2.5-VL-7B-Instruct & Ours            & 94.15 & 80.68 \\
                       & Ours + GPT-4o   & 94.26 & 79.63 \\
InternVL3-8B           & Ours            & 91.22 & 75.98 \\
                       & Ours + GPT-4o   & 92.69 & 76.36 \\ \bottomrule
\end{tabular}
}
\caption{Effect of different answer extractors on performance (MMBench).}
\label{tab:extraction-reliability}
\end{table}

\section{Conclusion}

% \textcolor{red}{We present a lightweight method for estimating question difficulty based on the hidden representations of LLMs. By modeling the generation process as a Markov chain and defining a value function over hidden representations, our approach enables efficient and accurate difficulty estimation without output generation. Experiments across diverse tasks show improved difficulty classification performance and enhanced inference efficiency when applied to adaptive reasoning.}

We propose a lightweight approach for estimating question difficulty by leveraging the hidden representations of LLMs. By modeling the generation process as a Markov chain and introducing a value function over hidden states, our method enables efficient and accurate difficulty estimation without requiring output generation. Experimental results across diverse tasks demonstrate that our approach improves difficulty classification performance and enhances inference efficiency when applied to adaptive reasoning.

\section{Limitations}

While our method avoids costly response sampling and preserves model generality, it requires access to token-level hidden representations from the target LLM, which may not be readily accessible in certain closed-source systems. Additionally, our approach currently focuses on single-turn inputs and may require adaptation for multi-turn or conversational settings. Exploring broader generalization to unseen domains and tasks remains future work.

\section{Acknowledgements}

We sincerely thank all the anonymous reviewers for their constructive feedback. This work was supported in part by the Shanghai Artificial Intelligence Laboratory, the National Natural Science Foundation of China (NSFC) under Grants 62372226, 62272215, and 62002159, and in part by the Fundamental Research Funds for the Central Universities.

\bibliography{emnlp}
\bibliographystyle{acl_natbib}

\clearpage
\appendix
\section{Appendix}
\label{sec:appendix}

\subsection{Datasets Details}
\label{dataset-details}

\subsubsection{MMBench}

MMBench~\cite{MMBench} is a multimodal dataset designed to evaluate the understanding capabilities of large language models. We randomly sample 50\% of the data as the training set, 45\% as the test set, and the remaining 5\% as the validation set for determining the threshold $\tau$.

\subsubsection{ScienceQA} 

ScienceQA~\cite{lu2022learn} is a multimodal dataset for science question answering, annotated with answers, lectures, and explanations. We use the official training, test, and validation splits provided by the dataset to determine the threshold $\tau$.

\subsubsection{MathVista}

MathVista~\cite{lu2024mathvista} is a dataset designed to combine challenges from diverse mathematical and visual reasoning tasks. We randomly sample 50\% of the data as the training set, 45\% as the test set, and the remaining 5\% as the validation set for determining the threshold $\tau$. 

\subsubsection{StrategyQA}

StrategyQA~\cite{geva2021did} is a dataset consisting of strategy questions, their decompositions, and supporting evidence paragraphs. We follow the official test split, and randomly sample 95\% of the official training data for training, with the remaining 5\% used as the validation set for determining the threshold $\tau$.

\subsubsection{gsm8k}

GSM8K~\cite{cobbe2021training} is a dataset of high-quality, linguistically diverse grade school math word problems. We follow the official test split, and randomly sample 95\% of the official training data for training, with the remaining 5\% used as the validation set for determining the threshold $\tau$. 

\subsubsection{commonsenseQA}

CommonsenseQA~\cite{talmor2018commonsenseqa} is a multiple-choice question answering dataset that requires diverse types of commonsense knowledge to predict the correct answer. We use the official training, test, and validation splits provided by the dataset to determine the threshold $\tau$.

\subsection{Baselines Details}
\label{baseline-details}

\subsubsection{prompt}
The method estimates problem difficulty by using a prompt to instruct the model to assess the difficulty of the input question itself. The prompt is shown in Table~\ref{tab:diff-prompt}.

\subsubsection{AG}
AG~\cite{lee2025semantic} is a method that estimates problem difficulty based on the consistency of the target model’s outputs. We implement Adaptive Gating for difficulty estimation following the original setup, which uses Chain-of-Thought (CoT) reasoning with $k = 10$ samples.

\subsubsection{LLMs-Ranking}
LLMs-Ranking~\cite{wang2024make} is a method that introduces an auxiliary LLM to directly assess the difficulty of a given problem. We reproduce its Difficulty Ranking and Problem Partition components for difficulty estimation. For Difficulty Ranking, we use Chain-of-Thought (CoT) sampling and follow the original settings by using a batch size $B = 8$ and the number of random split rounds $R = 5$. For Problem Partition, we set the pre-sample size $p = 4$ and the judge window size $k = 32$, consistent with the original implementation.

\subsubsection{HaluSearch-Gen}

HaluSearch-Gen~\cite{cheng2025think} is a training-based method that fine-tunes the LLM to equip it with the ability to assess problem difficulty.  We employ GPT-4o to generate reward data and fine-tune Qwen2.5-VL-7B-Instruct to equip the model with the ability to perceive question difficulty. The prompt used to generate reward data is shown as in Table~\ref{tab:gen-prompt}.

\subsubsection{HaluSearch-Critic}
HaluSearch-Critic\cite{cheng2025think} is a variant of HaluSearch-Gen that incorporates a critic signal into the training data to provide more explicit supervision for difficulty assessment. We employ GPT-4o to generate reward data and fine-tune Qwen2.5-VL-7B-Instruct to equip the model with the ability to perceive question difficulty. The prompt used to generate reward data is shown as in Table~\ref{tab:critic-prompt}.

\begin{table*}[h!]
\centering
\begin{tcolorbox}[colback=white, colframe=black!50, sharp corners=south, boxrule=0.4pt, width=0.95\linewidth]
You will be given a question between [Question begin] and [Question end]. And the image of this question will be provided.\\

Please answer the following:\\

Is this question difficult for you? Answer strictly with only "Yes" or "No". Do not provide any explanation.\\

\texttt{[Question begin]}\\
\texttt{{question}}\\
\texttt{[Question end]}\\
\end{tcolorbox}
\caption{Prompt used in baseline \textquotedblleft prompt\textquotedblright.}
\label{tab:diff-prompt}
\end{table*}

\begin{table*}[h!]
\centering
\begin{tcolorbox}[colback=white, colframe=black!50, sharp corners=south, boxrule=0.4pt, width=0.95\linewidth]
Please rate the difficulty of the following question for the model to answer correctly. The difficulty reflects how likely the model is to make mistakes, misunderstand, or fail to generate a complete and correct response. Use the provided correct and generated answers to guide your judgment.\\

There are five levels of question difficulty: 

1 - Very Easy: The question is straightforward, and the model is almost certain to answer it correctly. 

2 - Easy: The question is generally easy, though minor misunderstandings are possible. 

3 - Moderate: The model may partially struggle with this question, with some risk of mistakes or omissions. 

4 - Hard: The model is likely to make noticeable errors or fail to fully understand the question. 

5 - Very Hard: The question is highly challenging, and the model is very likely to answer it incorrectly or with significant flaws.\\

Only output the score (a number), do not give any explanation. Do not penalize for incomplete answers unless they indicate misunderstanding or error. Use the correct answer as a reference.\\

\texttt{[question begin]}\\
\texttt{\{question\}}\\
\texttt{[question end]}\\

Correct Answer: \texttt{\{correct\_answer\}}\\ 
Generated Answer: \texttt{\{generated\_answer\}}
\end{tcolorbox}
\caption{Prompt for generating reward data in HaluSearch-Gen.}
\label{tab:gen-prompt}
\end{table*}

\begin{table*}[h!]
\centering
\begin{tcolorbox}[colback=white, colframe=black!50, sharp corners=south, boxrule=0.4pt, width=0.95\linewidth]
Please rate the difficulty of the following question for the model to answer correctly.
The difficulty reflects how likely the model is to make mistakes, misunderstand, or fail to generate a complete and correct response. Use the provided correct and generated answers to guide your judgment.\\

There are five levels of question difficulty: 

1 - Very Easy: The question is straightforward, and the model is almost certain to answer it correctly.\\
2 - Easy: The question is generally easy, though minor misunderstandings are possible.\\
3 - Moderate: The model may partially struggle with this question, with some risk of mistakes or omissions.\\
4 - Hard: The model is likely to make noticeable errors or fail to fully understand the question.\\
5 - Very Hard: The question is highly challenging, and the model is very likely to answer it incorrectly or with significant flaws.\\

Give your explanation on the first line, and output the score (a number) on the second line. Do not penalize for incomplete answers unless they indicate misunderstanding or error. Use the correct answer as a reference.\\

\texttt{[question begin]}\\
\texttt{\{question\}}\\
\texttt{[question end]}\\

Correct Answer: \texttt{\{correct\_answer\}}\\ 
Generated Answer: \texttt{\{generated\_answer\}}
\end{tcolorbox}
\caption{Prompt for generating reward data in HaluSearch-Critic.}
\label{tab:critic-prompt}
\end{table*}

\clearpage

\subsection{Implementation Details}

\label{imply-details}

Our method is implemented using Python 3.9.21 and PyTorch 2.5.1, and runs on a single NVIDIA A100 GPU. The sampling temperature $T$ for LLMs is set as 0.5. For the two-layer fully connected neural network, the learning rate is set as $1\times 10^{-4}$.

\clearpage

\begin{figure*}[!t]
    \small
    \centering
    \includegraphics[width=0.95\textwidth,height=0.35\textheight]{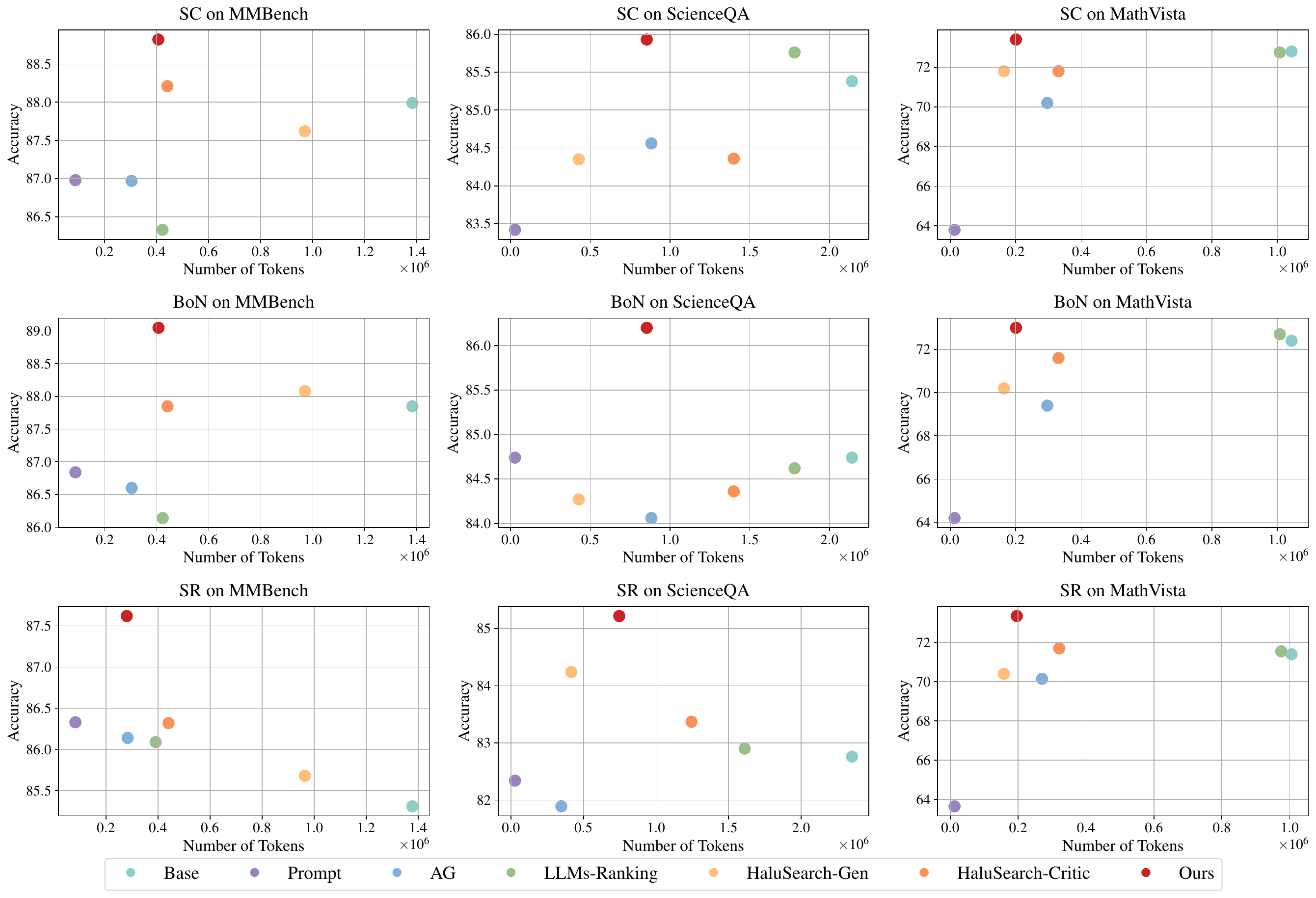}
    \caption{Performance comparison of difficulty-aware sampling methods across multiple datasets for Qwen2.5vl-7B-instruct. "SC" refers to Self-Consistency, "BoN" refers to the Best-of-N, and "SR" refers to Self-Refine. "Number of Tokens" represents the total number of output tokens generated on the test set. For SC and BoN, the budget $K$ is set to $5$, while for SR, the budget $K$ is set to $3$.}
    \label{fig:MMBench2}
\end{figure*}
\end{document}